\documentclass[11pt,a4paper]{article}
\usepackage{authblk}
\usepackage[hyperref]{naaclhlt2019}
\usepackage{times}
\usepackage{latexsym}

\aclfinalcopy 
\def\aclpaperid{1621} 

\usepackage{url}
\usepackage{graphicx}
\usepackage{amsmath}
\usepackage{amssymb}
\usepackage{pgfplots}
\pgfplotsset{compat=1.8}
\usepackage{mathtools}
\usepackage{tikz}
\usepackage{wrapfig}
\usepackage{caption}
\usepackage{subcaption}
\usepackage{cleveref}
\usepackage{tabularx}

\usepackage{algorithm}
\usepackage[noend]{algpseudocode}
\usepackage{relsize}
\usepackage{textcomp}
\usepackage{multirow}

\graphicspath{{./figures/}}

\title{News Article Teaser Tweets and How to Generate Them}

\author[1,2]{\bf Sanjeev Kumar Karn}
\author[2]{\bf Mark Buckley}
\author[2]{\bf Ulli Waltinger}
\author[1]{\bf Hinrich Sch\"{u}tze}
\affil[1]{Center for Information and Language Processing (CIS), LMU Munich}
\affil[2]{Machine Intelligence, Siemens CT, Munich, Germany}
\affil[1]{\tt skarn@cis.lmu.de}
\affil[2]{\tt \{sanjeev.kumar\_karn,mark.buckley,ulli.waltinger\}@siemens.com}

\date{}

\def\figref#1{Figure~\ref{fig:#1}}
\def\eqref#1{Eq.~\ref{eqn:#1}}

\def\tabref#1{Table~\ref{tab:#1}}
\def\tablabel#1{\label{tab:#1}\label{p:#1}}

\def\secref#1{Section~\ref{sec:#1}}

\newcounter{notecounter}

\newcommand{\enoteson}{\long\gdef\enote##1##2{{
\stepcounter{notecounter}
{\large\bf
\hspace{1cm}\arabic{notecounter} $<<<$ ##1: ##2
$>>>$\hspace{1cm}}}}}
\enoteson
\newcommand{\tabitem}{$\vcenter{\hbox{\tiny$\bullet$}}\,$}
\newcommand\boldblue[1]{\textcolor{blue}{\textbf{#1}}}
\newcommand\Tstrut{\rule{0pt}{2.6ex}}  

\def\algref#1{Algorithm.~\ref{alg:#1}}
\algdef{SE}[VARIABLES]{Variables}{EndVariables}
   {\algorithmicvariables}
   {\algorithmicend\ \algorithmicvariables}
\algnewcommand{\algorithmicvariables}{\textbf{global}}

\begin{document}

\maketitle
\begin{abstract}
In this work, we define the task of teaser generation and provide an evaluation benchmark and baseline systems for the process of generating teasers. A teaser is a short reading suggestion for an article that is illustrative and includes curiosity-arousing elements to entice potential readers to read particular news items. Teasers are one of the main vehicles for transmitting news to social media users. We compile a novel dataset of teasers by systematically accumulating tweets and selecting those that conform to the teaser definition. We have compared a number of neural abstractive architectures on the task of teaser generation and the overall best performing system is \newcite{P17-1099}'s seq2seq with pointer network.
\end{abstract}

\section{Introduction}
A considerable number of people get their news in some digital format.\footnote{http://www.journalism.org/2008/07/21/the-influence-of-the-web/} The trend has made many publishers and editors shift their focus to the web and experiment with new techniques to lure an Internet-savvy generation of readers to read their news stories. Therefore, there has been a noticeable increase in the sharing of short illustrative pieces of texts about the news on social media.




We define a ShortcutText as a short text (about 15 words or less) describing and pointing to a news article and whose purpose is to invite the recipient to read the article. A headline is a ShortcutText that optimizes the relevance of the story to its reader by including interesting and high news value content from the article \cite{DOR2003695}. Click-bait is a pejorative term for web content whose main goal is to make a user click an adjoining link by exploiting the information gap. According to the definition, a principal part of the headline is an extract of the article, thereby creating an impression of the upcoming story. However, click-bait, a ShortcutText, contains mostly elements that create anticipation, thereby making a reader click on the link; however, the reader comes to regret their decision when the story does not match the click-bait's impression \cite{BLOM201587}. Thus, click-bait provides a false impression (non-bona fide) and contains insufficient information (highly abstractive).

\begin{table}[ht!]
\begin{center}
\footnotesize
\begin{tabular}{l|c|c|c}
& bona-fide & teasing & abstractive
\\\hline
headline & yes & no & low\\
clickbait& no & yes & high\\
teaser & yes & yes & high 
\end{tabular}
\end{center}
\caption{\tablabel{shortcuts} The table shows three
  categories of ShortcutTexts and their properties}
\end{table}

We introduce the new concept of \emph{teaser} and define it as a ShortcutText devised by fusing curiosity-arousing elements with interesting facts from the article in a manner that concurrently creates a valid impression of an upcoming story and a sense of incompleteness, which motivates the audience to read the article. A teaser is one of the main vehicles for transmitting news on social media. \tabref{examples} shows some teasers from a popular newswire \textit{The Wall Street Journal}.

\begin{table}[h!]
\begin{center}
\footnotesize
\resizebox{0.99\linewidth}{!}{
\begin{tabular}{lp{0.95\linewidth}}
\hline\hline
Article & Global trade is in trouble, and investors dont seem to care. One of the ironies of the election of a fierce nationalist in the U.S. \ldots
\\
Headline & Steel Yourself for Trumps Anti-Trade Moves
\\
Teaser & Investors don't seem worried about a trade war. Could tariffs by Trump start one?\\\hline
Article & The U.S. Supreme Court on Monday partially revived President Donald Trumps executive order suspending travel from six countries \ldots
\\
Headline & High Court Says Travel Ban Not For Those With ‘Bona Fide’ Relationships
\\
Teaser & In a 'bona fide' relationship? You can visit the U.S.\\\hline
Article & Gan Liping pumped her bike across a busy street, racing to beat a crossing light before it turned red. She didnt make it. \ldots
\\
Headline & China’s All-Seeing Surveillance State Is Reading Its Citizens’ Faces	
\\
Teaser & China is monitoring its citizens very closely. Just ask jaywalkers.
\\
\hline\hline
\end{tabular}}
\end{center}
\caption{\tablabel{examples}The table contains tuples of news articles and their ShortcutTexts: headline and teaser. These tuples are from a popular newswire, \textit{The Wall Street Journal}.}
\end{table} 


We also introduce properties such as teasing, abstractive, and bona-fide, which not only differentiate teasers from other ShortcutTexts but also help in compiling a dataset for the study. Teasing indicates whether curiosity-arousing elements are included in the ShortcutText. Abstractive indicates whether a fair proportion of the ShortcutText is distilled out of the news article. Bona-fide answers whether the news story matches the impression created by the ShortcutText. \tabref{shortcuts} lists the common forms of the ShortcutTexts along with the presence or absence of the properties mentioned above.



In this study, we focus on teasers shared on Twitter\footnote{https://twitter.com/}, a social media platform whose role as a news conduit is rapidly increasing. An indicative tweet is a Twitter post containing a link to an external web page that is primarily composed of text. The presence of the URL in an indicative tweet signals that it functions to help users decide whether to read the article, and the short length confirms it as a ShortcutText like a headline or teaser. \newcite{DBLP:journals/eswa/LloretP13} made an early attempt at generating indicative tweets using off-the-shelf extractive summarization models, and graded the generated texts as informative but uninteresting. Additionally, \newcite{sidhaye-cheung:2015:EMNLP}'s analysis showed extractive summarization as an inappropriate method for generating such tweets as the overlaps between the tweets and the corresponding articles often 
are low. Our study shows that teasers, bona fide indicative tweets, do exhibit significant, though not complete, overlaps, and, therefore, are not appropriate for extractive but certainly for abstractive summarization. 

Our contributions:

1) To the best of our knowledge, this is the first attempt to compare different types of ShortcutTexts associated with a news article. Furthermore, we introduce a novel concept of a teaser, an amalgamation of article content and curiosity-arousing elements, used for broadcasting news on social media by a news publisher.  

2) We compiled a novel dataset to address the task of teaser generation. The dataset is a collection of news articles, ShortcutTexts (both teasers and headlines), and story-highlights. Unlike ShortcutText, a story-highlight is brief and includes self-contained sentences (about 25-40 words) that allow the recipient to gather information on news stories quickly. As all corpora based on news articles include only one of these short texts, our dataset provides the NLP community with  a unique opportunity for a joint study of the generation of many short texts.

3) We propose techniques like unigram overlap and domain relevance score to establish abstractivity and teasingness in the teasers. We also apply these techniques to headlines and compare the results with teasers. The comparison shows teasers are more abstractive than headlines. 

4) High abstractivity makes teaser generation a tougher task; however, we show seq2seq methods trained on such a corpus are quite effective. A comparison of different seq2seq methods for teaser generation shows a seq2seq combining two levels of vocabularies, source and corpus, is better than one using only the corpus level. Therefore, we set a strong baseline on the teaser generation task with a seq2seq model of \newcite{P17-1099}.

The remaining paper is structured as follows. In Section 2, we provide a detailed description of the data collection and analyses. In Section 3, we describe and discuss the experiments. In Section 4, we describe a user study of model-generated teasers. In Section 5, we discuss the related works. Section 6 concludes the study.

\section{Teaser Dataset}
Several linguistic patterns invoke curiosity, e.g., provocative questions and extremes for comparison. 
A retrieval of teasers from a social media platform using such patterns requires the formulation of a large number of complex rules as these patterns often involve many marker words and correspondingly many grammar rules. A computationally easy approach is to compile circulations from bona-fide agents involved in luring business on such media, and then filtering out those that don't comply with defined characteristics of a teaser; see \tabref{shortcuts}. We followed the latter approach and chose Twitter to conduct our study.

\subsection{Collection}
We identified the official Twitter accounts of English-language news publications that had tweeted a substantial number of times before the collection began; this removes a potential source of noise, namely indicative tweets by third-party accounts referencing the articles via their URL. We downloaded each new tweet from the accounts via Twitter's live streaming API. 
We limited the collection to indicative tweets and extracted the article text and associated metadata from the webpage using a general-purpose HTML parser for news websites.\footnote{https://github.com/codelucas/newspaper/} Overall, we collected approximately 1.4 million data items.

\subsection{Analysis}
We propose methods that evaluate teasingness and abstractivity in the teasers and verify them through analyses. We then combine those methods and devise a teaser recognition algorithm. Analyses are performed on lowercase, and stopwords-pruned texts.

\subsubsection{Extractivity}\label{sec:extractive}
For a given pair of strings, one is an extract of another
if it is a substring of it. Teasers
are abstractive, which we confirm by making sure that the
ShortcutText is not an extract of article
sentences. Additionally, a teaser of an article is designed
differently than the headline; therefore, they must be
independent of one other, i.e., non-extractive.

\subsubsection{Abstractivity}\label{sec:abstractive}
Abstractivity, a principle characteristic of the teaser, implies that the teaser should exhibit content overlap with its source, but not a full overlap.

We rely on \newcite{sidhaye-cheung:2015:EMNLP}'s method of computing the percentage match between two stemmed texts for grading  abstractivity. We obtain unigrams of the first, $\textit{X}_1$, and second text, $\textit{X}_2$, using function $uni(X)$ and compute the percentage match using \eqref{per_match}:
\begin{equation}
\footnotesize
perc\_match(\textit{X}_1, \textit{X}_2) = \frac{|uni(\textit{X}_1) \cap uni(\textit{X}_2)|}{|uni(\textit{X}_1)|}
\label{eqn:per_match}
\end{equation}

Given a ShortcutText and article, initially, a sequence of texts is obtained by sliding a window of size $p$ on the article sentences. Then, $perc\_match$ scores between the ShortcutText and sequence of texts are computed. A text with the highest score is selected as the prominent section for the ShortcutText in the article.

A full-overlap, i.e., $perc\_match$ of 1 is likely to be a case where the ShortcutText disseminates information of its prominent section. However, a non-overlap is very likely to be click-bait or noise. 
Thus, we filter out instances where the match score between a ShortcutText, potential teaser, and its prominent section is above 80\% or below 20\%. The intuition for the filtering is that the teasing words are likely to be absent from the prominent section, and an absence of a minimum of 2-3 words (often 20\%) is the easiest way to ascertain this fact. \tabref{abs_example} shows an example. Analogously, a presence of a minimum of 2-3 words from the source asserts that it is not click-bait or noise. 

We use the sliding window size, $p$, of 5,\footnote{Most of the prominent information is supposedly within a few leading sentences in the news articles due to the inverted pyramid news writing style.} and filter the data instances where the $perc\_match$ between the tweet and prominent section is lower than 0.2 or greater than 0.8. 

\begin{table}[t!]
\begin{center}
\footnotesize
\resizebox{0.99\linewidth}{!}{
\begin{tabular}{p{0.12\linewidth}|p{0.88\linewidth}}
\hline
Article & Diabetes medication, such as insulin, lowers blood sugar levels and $\ldots$. But experts have revealed a natural treatment for diabetes could be lurking in the garden. $\dots$ Fig leaves have properties that can help diabetes $\ldots$ . An additional remedy \dots
\\
headline & Diabetes treatment: Natural remedy made from fig leaves revealed
\\
Teaser & \boldblue{Would} you \boldblue{Adam} and \boldblue{Eve} it? Natural treatment for DIABETES could be \boldblue{growing} in your garden\\
\hline
\end{tabular}
}
\end{center}
\caption{\tablabel{abs_example} ShortcutTexts and their non-overlaps (bold).}
\end{table} 

\subsubsection{Teasingness}\label{sec:teasing}
Apart from abstractivity, teasers include words and phrases that tease and are are embedded by authors who often draw on their vast knowledge of style and 
vocabulary to devise teasers. A commonly recognizable pattern among them is the inclusion of unusual and interesting words in a given context, e.g., words like \textit{Adam} and \textit{Eve} in the example of \tabref{abs_example}.

The Pareto principle or the law of the vital few, 
states that the 2,000 of the most frequently used words in a domain cover about 80\% of the usual conversation texts 
\cite{nation_2001,newman05paretozipf}. At first glance,
filtering those abstractive ShortcutTexts that constitute
only frequent words should intuitively prune uninteresting
ones and save ones that are similar to the example in
\tabref{abs_example}. However, a closer look at the pruned
ShortcutTexts shows several interesting teasers with
substrings comprised of out-of-place frequent-words, e.g.,
\textit{Las Vegas gunman Stephen bought nearly \%\% guns
  legally. But none of the purchases set off any red flags},
with an interesting sentence fragment containing the  phrase \textit{red flags}. This suggests that the methodology that uses plain frequency of words is not sufficient for determining interesting information.
\begin{equation}
\footnotesize
\begin{split}
tf_{domain}(w, d) = \frac{|\textnormal{term}\, w \,\textnormal{in domain}\, d|}{|\textnormal{terms in domain}\, d|}\\
idf_{domain}(w) = log\,\frac{|\textnormal{domains}|}{|\textnormal{domains containing}\, w|}\\
dr(w, d) = tf_{domain}(w, d) \times idf_{domain}(w)
\end{split}
\label{eqn:dr_domain}
\end{equation} 

Thus, we look at unusualness at a level lower than the corpus. We rely on domain relevance ($dr$) \cite{schulder-hovy:2014:W14-23}, an adapted TF-IDF (term frequency inverse document frequency) metric that measures the impact of a word in a domain 
and, therefore, identifies unusual words in a specific domain, 
and is computed using \eqref{dr_domain}.

A word is assigned a very low $dr$ score if the word is either non-frequent in the domain and too frequent among other domains (unusualness) or non-frequent in all domains (rare); see \tabref{relevance}. As a very low $dr$ score corresponds to unusualness, a presence of very low $dr$ values among the non-overlapping words of the ShortcutText suggest a high likelihood of it being a teaser, and therefore, we compile them as teasers. However, the filtering requires a threshold  $dr$ value that defines anything lower than it as a very low $dr$. Also, computing $dr$ requires domain information of the text. 

\begin{table}[h!]
\begin{center}
\footnotesize
\resizebox{0.90\linewidth}{!}{
\begin{tabular}{l|l|l}
& $freq\_$Out & $\neg$ $freq\_$Out
\\\hline
$freq\_$IN & low & high \\
$\neg$ $freq\_$IN & very low & very low\\
\end{tabular}
}
\end{center}
\caption{\tablabel{relevance} The table shows $dr$ score range. IN and Out refer to in-domain and out-of-domain respectively.}
\end{table}

\begin{table}[!h]
\begin{center}
\resizebox{0.99\linewidth}{!}{
\begin{tabular}{p{0.2\linewidth}p{0.2\linewidth}p{0.2\linewidth}p{0.2\linewidth}p{0.2\linewidth}}
\hline
\boldblue{Would}&\boldblue{Adam}&\boldblue{Eve}& Natural& treatment\\
{0.104}&\textbf{0.0027}&\textbf{0.0025}&{0.025}&{0.016}\\
\end{tabular}
}
\resizebox{0.99\linewidth}{!}{
\begin{tabular}{p{0.05\linewidth}p{0.23\linewidth}p{0.23\linewidth}p{0.23\linewidth}p{0.23\linewidth}}
&DIABETES& could& \boldblue{growing}& garden\\
&{0.005}&{0.105}&{0.022}&{0.01}\\
\hline
\end{tabular}
}
\end{center}
\caption{\tablabel{example_tr_values}Teaser words and their $dr$ values. Non-overlaps are in bold blue.}
\end{table}


\begin{table*}[!ht]
\begin{center}
\footnotesize
\resizebox{\textwidth}{!}{
\begin{tabular}{p{0.11\linewidth}|p{0.11\linewidth}|p{0.11\linewidth}|p{0.11\linewidth}|p{0.11\linewidth}|p{0.11\linewidth}|p{0.11\linewidth}|p{0.11\linewidth}|p{0.125\linewidth}}
&Cluster 0& Cluster 1& Cluster 2& Cluster 3& Cluster 4& Cluster 5& Cluster 6& Cluster 7
\\\hline
Keywords&
politics, UK, world, Brexit, Europe, UK Politics, Theresa May&
Trump, United States, election, immigration, White House&
culture, travel, home and garden, food and beverage&
entertainment, celebrities, movies, concerts, Netflix&
Europe, Russia, North Korea, Diplomacy, Conflicts&
shooting, police, murder, killed, dead, fire, suspect, crash&
Corporate, business, Company, automotive, Equities&
Sport, Football, Premier League, NFL, Dallas Cowboys, Rugby\\
\hline
Avg. size (words)&763&842&526&838&886&651&791&741
\end{tabular}}
\end{center}
\caption{\tablabel{domain_cluster}The table shows clusters of domains and corresponding frequent-keywords and average article size (words) in them.}
\end{table*}

\subsubsection*{Obtaining Domains}
We make use of articles and their keywords to determine domains. Keywords are meta-information available for a subset of corpus instances. We rely on Doc2vec \cite{pmlr-v32-le14} for obtaining the representations for the articles and cluster these representations by K-Means clustering \cite{hartigan1979algorithm}. 

We rely on elbow criterion and uniformity among keywords in the clusters to determine the number of clusters. The uniformity is validated by manual inspection of 100 most-frequent keywords. Clustering the corpus into eight domains resulted in the final abrupt decrease of the Sum of Squared Error (SSE) as well as uniformly distributed keyword sets. See \tabref{domain_cluster} for domain-wise keywords and other statistics.




\subsubsection*{Selecting a Threshold}
We use the domain information and compute $dr$ values of potential teaser texts in the corpus. \tabref{example_tr_values} shows nonstop words and $dr$ scores for \tabref{abs_example} example. Evidently, unusual words have very low $dr$ scores (bold values).

To determine an appropriate threshold, we design an unsupervised methodology based on the Pareto principle. The cue remains the same, i.e., a right threshold will filter only the teasers, and the non-overlapping words in them are less likely to be frequent words. 

\begin{figure}[!ht]
\resizebox{0.99\linewidth}{!}{
\includegraphics[width=1.cm, height=1.cm,keepaspectratio,clip,trim={0.20cm 0 0 0}]{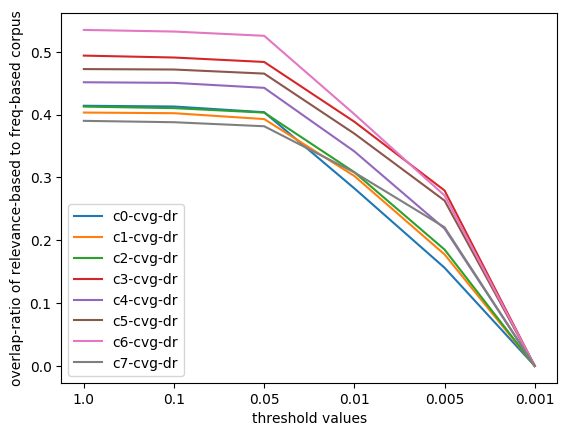}
}
\caption{Overlap-ratio of frequency-based and threshold-based filtered teasers for  domains (c\#). }
\label{fig:drelevance-pareto-filter}
\end{figure}

Thus, we define a range of possible threshold values, and for each value, we compile a corpus of teasers where a non-overlapping word has $dr$ below it. Meanwhile, we also compile sets of most-frequent words that cover 80\% of the total word occurrences in all 8 domains (sizes $\approx$ 2000). Then, we determine the ratio of the teasers that have their non-overlapping words completely overlapping the frequent word sets. Finally, we select a value which has the least overlap as the threshold; see \figref{drelevance-pareto-filter}. We chose 0.005 as it is the boundary below which there is no overlap. We apply this value to abstractive ShortcutTexts and obtain a teaser corpus.

\subsection{Teaser Recognition Algorithm}
We combine the above three methodologies and devise a teaser recognition algorithm; see \algref{teaser_xtract}.

We use notations like uppercase bold for a matrix, lowercase italic for a variable and uppercase italic for an array. A data instance in the corpus has an article $\textit{A}$, headline $\textit{H}$, tweet $\textit{T}$, and domain $\textit{d}$. An article, $\textit{A}$, has a sequence of sentences, $\textit{S} = \langle\textit{S}_1,\ldots,\textit{S}_{|\mathit{A}|}\rangle$, and each sentence, $\textit{S}_{i}$, has a sequence of words, $\langle\textit{w}_{1},\ldots,\textit{w}_{|\mathit{S}_{i}|}\rangle$. $\textsc{window}$ takes a sequence of sentences, $\textit{S}$, and returns a sequence of texts, $\textit{Z}$, of size $\frac{|\mathit{S}|-\textit{p}}{q}+1$, where $\textit{p}$ and $\textit{q}$ are window size and sliding step respectively. The domain-wise $dr$ values for words in the vocabulary, $\mathit{U}$, is stacked into a matrix, $\textbf{D}$. $\textsc{Is\_Teaser}$ takes $\textbf{D}$ and items of a data instance, and determines whether its tweet, $\textit{T}$, is a teaser.

\begin{algorithm}[!h]
\footnotesize
\caption{Teaser Recognition}\label{alg:teaser_xtract}
\begin{algorithmic}[1]
\Procedure{Is\_Teaser}{$\textit{A}, \textit{H}, \textit{T}, \textit{d}, \textbf{D}$}
\If {($\textit{T}$ in $\textit{H}$) Or ($\textit{H}$ in $\textit{T}$)} 
\State \Return False \Comment{sub-string match}
\EndIf
\For {$\textit{S}$ in $\textit{A}$}
\If {($\textit{T}$ in $\textit{S}$) Or ($\textit{S}$ in $\textit{T}$)} 
\State \Return False \Comment{sub-string match}
\EndIf
\EndFor

\State $\textit{Z} \gets \textsc{window}(\textit{A}, p=5,q=1)$
\State $\textit{V} \gets$ Array()
\For {$i$ = 1 to $|\textit{Z}|$}
    \State $\textit{V}.\textsc{Add}(perc\_match(\textit{Z}[i]$, $\textit{T})$) \Comment{see \eqref{per_match}}
\EndFor

\If {$\textit{max}$($\textit{V}$) $> 0.8$ Or $\textit{max}$($\textit{V}$) $<0.2$}:
    \State \Return False   \Comment{abstractivity}
\EndIf
\State $\hat{\textit{Y}} \gets \textit{Z}$[$\textit{max}$($\textit{V})$]\Comment{prominent section}
\State $\textit{T}^{\prime} \gets \hat{\textit{Y}}$\textbackslash$ \textit{T}$\Comment{non-overlap}
\State $\textit{L} \gets \textbf{D}$[$d$;$\textit{T}^{\prime}$]\Comment{indexing}
\If {any ($\textit{L} < 0.005$)} 
	\State \Return True \Comment{teasingness}
\EndIf 

\State \Return False       
\EndProcedure
\end{algorithmic}
\end{algorithm}

Overall, in \algref{teaser_xtract}, steps 2 to 6 checks Extractivity, steps 7 to 12 checks Abstractivity, and steps 13 to 17 checks Teasingness. \tabref{trecog_filter} shows the percentage distribution of the total data points that are pruned by each of those analyses. Finally, we compile the remaining 23\% data points, i.e., 330k as a teaser corpus. 

\begin{table}[ht!]
\begin{center}
\resizebox{0.90\linewidth}{!}{
\begin{tabular}{p{0.30\linewidth}|>{\centering\arraybackslash}p{0.30\linewidth}|>{\centering\arraybackslash}p{0.40\linewidth}}
\multicolumn{2}{c|}{Analysis} & \% pruned
\\\hline
\multirow{2}*{Extractivity} & wrt headline & 37\% 
\\
&wrt article& 5\% 
\\\hline
\multicolumn{2}{l|}{Abstractivity} & 22\% 
\\\hline
\multicolumn{2}{l|}{Teasingness} & 13\% 
\\\hline
\end{tabular}
}
\end{center}
\caption{\tablabel{trecog_filter} The table shows different analyses performed in \algref{teaser_xtract} and the corresponding approximate percentage of data points pruned using them. ``wrt'' = with respect to. }
\end{table}


 
\begin{figure}[!hb]
\centering
\begin{subfigure}{.5\textwidth}
  \centering
  \includegraphics[width=5.2cm, height=4.cm,keepaspectratio,clip,trim={0.32cm 0 0 0}]{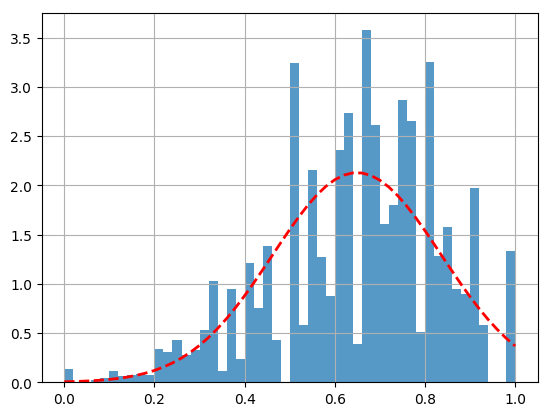}  
  \caption{\label{fig:perc_match_hist}teasers ($t1$) and articles ($t2$).}
\end{subfigure}
\begin{subfigure}{.5\textwidth}
  \centering
  \includegraphics[width=5.2cm, height=4.cm,keepaspectratio,clip,trim={0.32cm 0 0 0}]{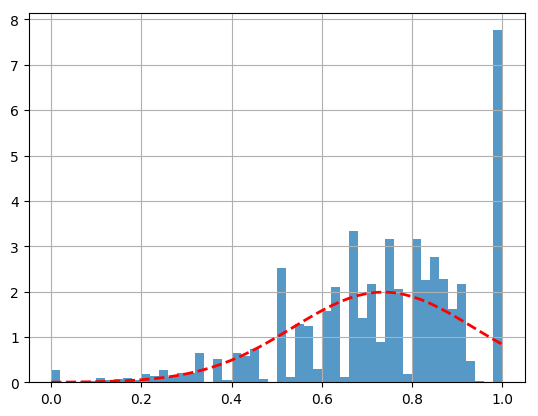}
  \caption{\label{fig:perc_match_hist_hdl}headlines ($t1$) and articles ($t2$)}
\end{subfigure}
\caption{Histogram of unigram overlaps obtained using \eqref{per_match}. The histograms are normalized, i.e., the area under the curve ($\sum$bin-height$\times$bin-width) sum up to one.}
\end{figure}

\subsection{Comparing ShortcutTexts}
The two ShortcutTexts, headline and teaser, have distinct conveyance mediums and therefore are designed differently, e.g., mean lengths of 10 and 14 respectively. However, abstractivity is also presumed for the headline. Therefore, we conduct additional overlap-based studies to understand the differences in the abstractive property between them. We compute and plot the distribution of the overlaps between teasers ($\textit{T}_1$) and articles ($\textit{T}_2$), and one between headlines ($\textit{T}_1$) and articles ($\textit{T}_2$); see \figref{perc_match_hist} and \figref{perc_match_hist_hdl} for respective plot. Clearly, compared to the teaser, headline distribution is left-skewed (mean 74\% and std 20\%), and thereby implies that headlines have a lesser abstractive value than teasers. 

Further, a review of a few instances of headline-article instances with lesser than 60\% overlap reveals cases of noisy headlines or HTML-parse failures; therefore, in a typical scenario a headline with a size of 10 words takes nearly all of its content ($\approx$80\%) from the source while a teaser of size 14 has sufficient non-extractive contents ($\approx$32\%). See \tabref{abs_example} for an example.



\section{Experiments}
\subsection{Models}
We experiment with two state-of-the-art neural abstractive summarization techniques, attentive seq2seq \cite{DBLP:journals/corr/BahdanauCB14} and pointer seq2seq \cite{P17-1099}, for teaser generation. Attentive seq2seq learns to generate a target with words from a fixed vocabulary, while pointer seq2seq uses a flexible vocabulary, which is augmented with words from the source delivered through the pointer network. We refer to the individual papers for further details. 



\textbf{Evaluation Metrics:} 
Studies on text-summarization evaluate their system using Rouge; 
therefore, we report Rouge-1 (unigram), Rouge-2 (bigram), and Rouge-L (longest-common substring) as the quantitative evaluation of models on our corpus.


\textbf{Parameters:}
We initialized all weights, including word embeddings, with
a random uniform distribution with mean 0 and standard
deviation 0.1. The embedding vectors are of dimension
100. All hidden states of encoder and decoder in the seq2seq
models are set to dimension 200. We pad short sequences with
a special symbol $\langle PAD\rangle$. We use Adam 
with initial learning rate .0007 and batch size 32 for training. Texts are lowercased and numbers are replaced by the special symbol $\%$. The token length for the source is limited to 100 and target sequence to 25. The teaser baseline experiments and headline generation use vocabulary size of 20000.

\subsection{Baseline Setting}\label{sec:baseline}
As we reimplemented \cite{DBLP:journals/corr/BahdanauCB14} and \cite{P17-1099} models, we initially evaluate them on a standard task of headline generation.\footnote{codes for collection, analyses and experiments: \url{https://github.com/sanjeevkrn/teaser_collect.git} and \url{https://github.com/sanjeevkrn/teaser_generate.git}} We use popular headline generation corpus, Gigaword \cite{napoles2012annotated}, with 3.8M training examples. We fetched the test set from \newcite{DBLP:conf/emnlp/RushCW15} and report the results on it. 
The results are compared with the state-of-the-art headline generation methods like  Nallapati et al. \cite{DBLP:conf/conll/NallapatiZSGX16}, 
ABS \cite{DBLP:conf/emnlp/RushCW15}, 
ABS+ \cite{DBLP:conf/emnlp/RushCW15}, and RAS-Elman \cite{DBLP:conf/naacl/ChopraAR16}. 
Since our aim for this experiment is to demonstrate the strength of the models, we limit the model parameters to the extent that we produce comparable results in less computation time. \tabref{duc_perf_comp} compares performances of seq2seq and seq2seq\_pointer models with other state-of-the-art methods. The results indicate that the implementations have performance competitive with other state-of-the-art methods.

\def\evalsep{0.20cm}
\def\perlevelsep{0.125cm}

\begin{table}[t!]
\begin{center}
\footnotesize
\begin{tabular}{l|l@{\hspace{\evalsep}}l@{\hspace{\evalsep}}l}
 &  Rouge-1 &  Rouge-2 &  Rouge-L \\ \hline
ABS & 29.55 & 11.32 & 26.42 \\
ABS+ & 29.76 & 11.88 & 26.96 \\
RAS-Elman & 33.78 & \bf 15.97 & 31.15\\
Nallapati et al.& 32.67 & 15.59 & 30.64\\
seq2seq & 31.21 & 12.96  & 28.87\\
seq2seq\_point & \bf 34.81 & 15.59 & \bf 32.05\\
\end{tabular}
\end{center}
\caption{\tablabel{duc_perf_comp} Rouge scores on the standard task of Headline Generation (Gigaword). seq2seq and seq2seq\_point are reimplementations of \newcite{DBLP:journals/corr/BahdanauCB14} and \newcite{P17-1099} respectively.}
\end{table}

\begin{table}[ht!]
\begin{center}
\footnotesize
\begin{tabular}
{l|
c@{\hspace{\perlevelsep}}c@{\hspace{\perlevelsep}}c|}
&\multicolumn{3}{c|}{\bf Validation}
\\
& Rouge-1 &  Rouge-2 &  Rouge-L\\ 
\hline
seq2seq & 15.77& 03.52& 13.53\\
seq2seq\_point & \bf 21.57 & \bf 07.03& \bf 18.64
\\ 
\Tstrut &\multicolumn{3}{c|}{\bf Test}
\\ 
\hline
seq2seq & 15.26	& 03.38	& 13.15\\
seq2seq\_point & \bf 21.05 & \bf 07.11 & \bf 18.49
\end{tabular}
\end{center}
\caption{\tablabel{twgen_perf_comp} Rouge F1 scores for seq2seq model and seq2seq\_point models on the teaser task.}
\end{table}


These models are then trained and evaluated on the teaser
corpus obtained using \algref{teaser_xtract} that initially has 330k instances. We then sample 255k instances that have all associated short texts in them. The sampled corpus is split into non-overlapping
250k, 2k and 2k sets for training, validation, and testing,
respectively.
The split is constructed such that training, validation and
test sets have equal representation of all eight domains.
Any instances that describe events that were also described
in training are removed from
validation and test sets; thus, instances encountered in
validation / test are quite distinct from instances
encountered in training.
Models were selected
based on their performance on the validation
set. \tabref{twgen_perf_comp} shows the performance
comparison. Clearly, seq2seq\_point performs better than
seq2seq due to the boost in the recall gained by copying
source words through the pointer network.


\begin{table}[ht!]
\begin{center}
\footnotesize
\begin{tabular}
{l|
c@{\hspace{\perlevelsep}}c@{\hspace{\perlevelsep}}c|}
&\multicolumn{3}{c|}{\bf Teaser}
\\
& Rouge-1 &  Rouge-2 &  Rouge-L\\ 
\hline
seq2seq & 15.26	& 03.38	& 13.15\\
seq2seq\_point & \bf 21.05 & \bf 07.11 & \bf 18.49
\\ 
\Tstrut &\multicolumn{3}{c|}{\bf Headline}
\\ 
\hline
seq2seq & 18.52 & 05.34 & 16.74\\
seq2seq\_point & \bf 23.83 & \bf 08.73 & \bf 21.68
\\
\Tstrut &\multicolumn{3}{c|}{\bf Highlights}
\\ 
\hline
seq2seq & 31.18 & 17.57 & 27.30 \\
seq2seq\_point & \bf 35.92& \bf 22.44 & \bf 31.53
\end{tabular}
\end{center}
\caption{\tablabel{shorttext_perf_comp} Rouge F1 scores for
  seq2seq model and seq2seq\_point models on the teaser,
  headline and highlights generation task.}
\end{table}

Additionally, models are also trained and evaluated on the other short texts that are available in the novel corpus: headlines (also a ShortcutText) and story-highlights. All the model parameters remain the same except the generation size, which depends on the short text average size, e.g., 35 for highlights. \tabref{shorttext_perf_comp} compares the performance on the test data. Clearly, seq2seq\_point performs better than seq2seq for all the types of short texts. Additionally, the change in the rouge scores with the change of dataset, i.e., Teaser$<$Headline$<$Highlights, also corresponds to the level of distillation of source information in them.   

\begin{table}[ht!]
\begin{center}
\footnotesize
\resizebox{0.99\linewidth}{!}{
\begin{tabular}{p{0.13\linewidth}|p{0.87\linewidth}}
\hline
Article & Millions of disease carrying mosquitoes are to be freed in a well-meaning bid $\ldots$. The lab-grown versions are infected with a disease which prevents natural mosquito $\dots$. But some activists fear the disease could transfer to humans ultimately making all human males sterile $\ldots$ . Despite claims it is safe for humans , there are also some concerns $\dots$ rendering humans unable to breed  \dots
\\
\Tstrut &\multicolumn{1}{c}{\bf Ground-Truth}\\
\hline
\Tstrut Headline & \boldblue{SHOCK} CLAIM: Lab created super-mosquitos released into wild could 'make all men infertile'
\\
\mbox{Highlight} & A \boldblue{NEW} lab-designed mosquito being released into the wild could \boldblue{end} the human race by making men sterile, it was claimed \boldblue{today}.\\
Teaser & \boldblue{PLAYING} \boldblue{GOD}? Millions of lab-grown diseased mosquitoes to be released into \boldblue{wild}\\
\Tstrut &\multicolumn{1}{c}{\bf Generated}\\
\hline
\Tstrut Headline & millions of mosquitoes could be freed , study \boldblue{says}
\\
\mbox{Highlight} & millions of disease carrying mosquitoes are to be freed in a well-meaning bid to decimate natural populations of the malaria-spreading insects .\\
Teaser & activists fear the disease could be freed - but there 's a \boldblue{catch}\\
\hline
\end{tabular}
}
\end{center}
\caption{\tablabel{gen_example} seq2seq\_pointer generated examples. Non-overlapping words are in bold blue. More examples in Supplementary A.2.}
\end{table} 

\tabref{gen_example} shows an example of a data instance in the corpus and seq2seq\_point model generations. Among generations, only headline and teaser have non-overlapping words; however, the headline non-overlap, \textit{says}, is a frequent word with a high $dr$ (0.11) while the teaser non-overlap, \textit{catch}, is a domain-wise non-frequent one, and therefore, has a very low $dr$ (0.006). 

Further, the teaser is the most detached from the core news information among the three generations, while still being relevant. The generated highlight is extractive, and this is a reason for relatively high Rouge scores for highlights (see \tabref{shorttext_perf_comp}). Rouge is an overlap-based measure and, therefore, is inclined towards extractive datasets.

\subsection{Impact of very low $dr$}
We performed additional experiments to study the impact that can be generated using the domain relevance ($dr$). 
All the settings are kept intact as in \secref{baseline} except the training corpus; this is changed by increasing the proportion of very low $dr$ ($<$0.005) terms in the teasers. New models are trained using equal size training instances sampled out of the revised corpora.
 
\begin{figure}[h]
\begin{center}
\includegraphics[width=7.2cm, height=7.2cm,keepaspectratio,clip,trim={0.2cm 0 0 0}]{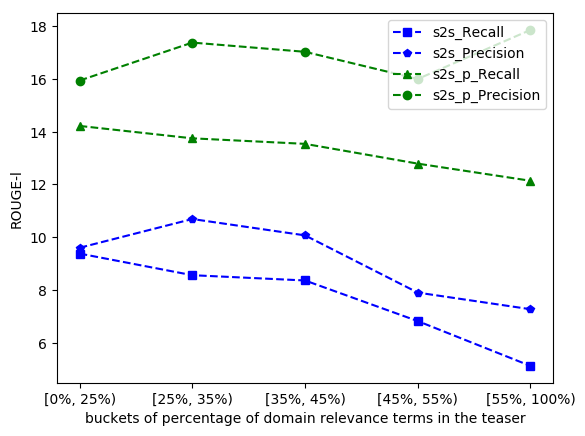}
\end{center}
\caption{\label{fig:perc_tr_rouge1}Variation in ROUGE-l on increasing the proportion of domain-relevant terms in the teasers. Models trained on 40k sampled instances.}
\end{figure}

A bucketing of very low $dr$ percentages into [0\%, 25\%), [25\%, 35\%), [35\%, 45\%), [45\%, 55\%) and [55\%, 100\%) divides the corpus into approximately equal sizes. Also, the mean and standard deviation of teaser-article overlap ratio is nearly equal in all the buckets, i.e., 0.559$\pm$0.148, 0.559$\pm$0.146, 0.564$\pm$0.146, 0.566$\pm$0.142, 0.566$\pm$0.146, respectively. Thus, the range of buckets corresponds to a range in the percentage of uncommon words. We evaluate the precision and recall of the models. Recall ($|overlap|$/$|ground$-$truth|$) estimates the model capacity in recovering the ground-truth content, while precision ($|overlap|$/$|generation|$) estimates the relevancy in the generation.

As shown in \figref{perc_tr_rouge1}, the test recall for both models decreases with the increase in uncommon words in their training. 
An increase in the proportion of uncommon words makes the models also generate uncommon words, which are not likely to match the ground-truth, thereby reducing the recall. 
However, in extreme cases, i.e., [45\%, 100\%), not only training teasers get slightly shorter but also a relatively large proportion of out-of-vocabulary (UNK) is introduced in them, and thereby in the generations. The UNK appears for novel informative words, which are rare words with a very low $dr$ as well (see \tabref{relevance}). Unlike seq2seq, seq2seq\_pointer recovers those from the source using pointer network and thus doesn't suffer an abrupt drop in the scores. 

Further, the precision scores in extreme cases have a slightly different trend than recall scores, and this is due to shorter generations, which supports precision, but is irrelevant for recall.

\begin{table}[t!]
\begin{center}
\footnotesize
\resizebox{0.99\linewidth}{!}{
\begin{tabular}{p{0.95\linewidth}}
\hline
\tabitem pres . trump lashed out on twitter at the hosts of `` msnbcs morning ''
\\
\tabitem migration agency says more than \%\% people drowned and presumed dead in myanmar to bangladesh\\
\tabitem computer glitch led to google to be dramatically undervalued this morning
\\
\tabitem alt-right activist jason kessler says he was swarmed by a group of charlottesville\\
\tabitem of identical triplets who beat the incredible odds of \%\%\% million to survive\\
\hline
\end{tabular}
}
\end{center}
\caption{\tablabel{generated_abs_teasers}Seq2seq\_point generated teasers used in the survey-based study. More examples in Supplementary A.2.}
\end{table}

\section{Human Evaluation}

\def\perlevelsep{0.125cm}
\begin{table}[t!]
\begin{center}
\footnotesize
\begin{tabular}
{l|
c@{\hspace{\perlevelsep}}c@{\hspace{\perlevelsep}}|c@{\hspace{\perlevelsep}}c}
&\multicolumn{2}{c|}{On Twitter}&\multicolumn{2}{c}{Stimulating}
\\
& Mean & Std & Mean & Std\\ 
\hline
ground-truth & 0.660 & 0.064 & 0.621 & 0.079\\
seq2seq\_point teaser & 0.588 & 0.078 & 0.559 & 0.089\\
baseline & 0.476 & 0.127 & 0.501 & 0.111\\
\end{tabular}
\end{center}
\caption{\tablabel{quali_results2} Mean and standard
  deviation of likelihood of being social-media text and
  stimulating for users to read. Baseline = lead sentences}
\end{table}
The quantitative evaluations show that state-of-the-art models perform moderately on the novel task. This is mostly due to deficiencies of  Rouge, which fails to reward heterogeneous contents. We took a closer look at some of the generated examples, see \tabref{generated_abs_teasers}, and observed frequent cases where the generation suffered from the typical seq2seq issues, e.g., repetition of words; however, there are also cases where generation is more distinctive than ground-truth and is well formed too. Thus, we carried out a small user study to understand the quality of the generated teasers; however, we only selected non-repeating and non-UNK generations to anonymize the source. The participants in the user study are undergraduate or graduate students with some computer science background and familiarity with social media platforms. Additionally, all the participants have used or have been using twitter.

We assembled a set of texts by randomly sampling 40
seq2seq\_point teasers, 40 ground-truth teasers, and 40 lead
sentences (baseline), and also established equal
representation of the domains. We then assigned 72 sentences
(3 per domain per category) to ten participants and asked
them to rate texts for two questions: 1) How likely is it
that the text
is shared on Twitter for a news story by a news organization?
and 2) How likely is it that the text makes a reader want to read
the story? The first question helps us recognize the
participant's understanding of the teasers, as an informed
reader will rate a ground-truth significantly higher than
the baseline, and 8 of them recognized it correctly, and
their ratings are selected for the evaluation. The second
question provides a cue as to the model capacity in generating
teasing texts by learning interesting aspects present in the
teaser corpus.

The annotators rated samples on a scale of 1 to 5; however, we normalized the ratings to avoid the influence of annotators having different rating personalities. The results, summarized in \tabref{quali_results2}, show that the human written teasers are most likely to be recognized as social media texts due to their style, which is distinct from the lead sentence; the model trained on such teasers closely follows it. Similarly, human written teasers are good at stimulating readers to read a story compared to the lead sentence and the generated teasers.

\section{Related Work}
There are two kinds of summarization: abstractive and extractive. In abstractive summarization, the model utilizes a corpus-level vocabulary and generates novel sentences as the summary, while extractive models extract or rearrange the source words as the summary. Abstractive models based on neural sequence-to-sequence (seq2seq) \cite{DBLP:conf/emnlp/RushCW15} proved to generate summaries with higher Rouge scores than the feature-based abstractive models. The integration of attention into seq2seq \cite{DBLP:journals/corr/BahdanauCB14} led to further advancement of abstractive summarization \cite{DBLP:conf/conll/NallapatiZSGX16,DBLP:conf/naacl/ChopraAR16,P17-1099}.

There are studies utilizing cross-media correlation like coupling newswire with microblogs; however, most of them involve improving tasks on newswire by utilizing complementary information from microblogs, e.g., improving news article summarization using tweets \cite{DBLP:conf/cikm/GaoLD12,DBLP:conf/coling/WeiG14}, 
generating event summaries through comments \cite{N15-1112},
etc. There is very limited work on using newswire and
generating microblogs, e.g., article tweet generation
\cite{DBLP:journals/eswa/LloretP13} and indicative tweet
generation
\cite{sidhaye-cheung:2015:EMNLP}. \newcite{DBLP:journals/eswa/LloretP13}
observed that off-the-shelf extractive models produce
summaries that have  high quantitative scores, but that are not interesting enough. 
Similarly, \newcite{sidhaye-cheung:2015:EMNLP}'s analysis of indicative tweets shows the narrow overlap between such tweets and their source limits the application of an extractive method for generating them. Our controlled compilation of such tweets shows a mean percentage match of 68.3\% (std: 16\%) with its source. These analyses strongly suggest that indicative tweets are not regular information-disseminating short texts. Also, the mixed nature of such texts suggests an abstractive, rather than extractive study. 

Most abstractive summarization systems use a popular
dataset, CNN/DailyMail\cite{napoles2012annotated}, that
includes news articles and story highlights to train and
test their performance. However, story highlights are brief
and self-contained sentences (about 25-40 words) that allow
the recipient to quickly gather information on news stories;
it is largely extractive
\cite{woodsend-lapata:2010:ACL}. Our novel corpus includes
not only extractive short texts (i.e., story-highlights) and
nearly extractive (i.e., headlines), but also very
abstractive teasers, and therefore is a challenging and more appropriate dataset to measure abstractive systems.

\section{Conclusion}
We defined a novel concept of a teaser, a ShortcutText amalgamating interesting facts from the news article and teasing elements. We compiled a novel dataset that includes  all of the short texts that are associated with news articles. We identified properties like abstractive, teasing, and bona-fide that assist in comparing a teaser with the other forms of short texts. We illustrated techniques to control these properties in teasers and verified their impact through experiments. An overlap-based comparative study of headlines and teasers shows teasers as abstractive while headlines as nearly extractive. Thus, we performed neural abstractive summarization studies on teasers and set a strong benchmark on the novel task of teaser generation.

\section*{Acknowledgments}
We thank Siemens CT members and the anonymous reviewers for valuable feedback. This research was supported by Bundeswirtschaftsministerium (\url{bmwi.de}),
grant 01MD15010A (Smart Data Web).

\bibliography{twgen}
\bibliographystyle{acl_natbib}

%
%
%
%
%
%
%
%
%

\appendix

\section{Supplemental Material}
\label{sec:supplemental}


\subsection{List of Twitter accounts}
The following is the list of Twitter accounts from which data was collected.

\begin{tabular}{ll}
Account ID & Account name\\
\hline
        759251  & CNN\\
        807095  & nytimes\\
        35773039  & theatlantic\\
        14677919  & newyorker\\
        14511951  & HuffingtonPost\\
        1367531  &FoxNews\\
        28785486  & ABC\\
        14173315  &NBCNews\\
        2467791  &washingtonpost\\
        14293310  &TIME\\
        2884771  &Newsweek\\
        15754281  &USATODAY\\
        16273831  &VOANews\\
        3108351  &WSJ\\
        14192680  &NOLAnews\\
        15012486  &CBSNews\\
        12811952  &Suntimes\\
        14304462  &TB\_Times\\
        8940342  &HoustonChron\\
        16664681  &latimes\\
        14221917  &phillydotcom\\
        14179819  &njdotcom\\
        15679641  &dallasnews\\
        4170491  &ajc\\
        \hline
        \end{tabular}
        
        \begin{tabular}{ll}
        \hline
        6577642  &usnews\\
        1652541  &reuters\\
        12811952  &suntimes\\
        7313362  &chicagotribune\\
        8861182  &newsday\\
        17820493  &ocregister\\
        11877492  &starledger\\
        14267944  &clevelanddotcom\\
        14495726  &phillyinquirer\\
        17348525  &startribune\\
        87818409  & guardian\\
        15084853  & IrishTimes\\
        15438913  & mailonline\\
        5988062  & theeconomist\\
        17680050  & thescotsman\\
        16973333  & independent\\
        4970411  & ajenglish\\
        \hline
        \end{tabular}

\subsection{Results}

\begin{table}[ht!]
\begin{center}
\footnotesize
\resizebox{0.99\linewidth}{!}{
\begin{tabular}{p{0.13\linewidth}|p{0.87\linewidth}}
\hline
&Sir Robert Fellowes , the Queen 's private secretary , was on the verge of making the extraordinary request $\ldots$. But he was persuaded to back off by fellow courtiers and the party went ahead as planned putting Camilla $\dots$. a visible and acknowledged part of the Prince 's life . A new book , The Duchess : The Untold Story by Penny Junor , makes sensational claims about Prince Charles $\ldots$ . furious about the birthday party even though by this stage she was fairly relaxed about Charles 's relationship \dots\\
\Tstrut &\multicolumn{1}{c}{\bf Ground-Truth}\\
\hline
\Tstrut Headline & Princess Diana latest: Queen aide planned to end Charles affair with Camilla amid rage
\\
\mbox{Highlight} & PRINCE Charles faced being told by the Queen to end his relationship with Camilla after Princess Diana erupted with fury over a lavish party for the Duchess-to-be , it is claimed .\\
Teaser & \boldblue{Royal} \boldblue{intervention} threatened Charles and Camilla's affair after Diana's \boldblue{fury} at \boldblue{posh} bash\\
\Tstrut &\multicolumn{1}{c}{\bf Generated}\\
\hline
\Tstrut Headline & duchess of the queen 's private secretary robert fellowes
\\
\mbox{Highlight} & sir robert fellowes , the queen 's private secretary , was on the verge of making the extraordinary request of her majesty .\\
Teaser & penny junor \boldblue{reveals} why she was on the \boldblue{brink} of making the queen 's life\\
\hline
\end{tabular}
}
\end{center}
\caption{seq2seq\_pointer generated examples. Non-overlapping words are in bold blue.}
\end{table} 

\begin{table}[ht!]
\begin{center}
\footnotesize
\resizebox{0.99\linewidth}{!}{
\begin{tabular}{p{0.13\linewidth}|p{0.87\linewidth}}
\hline
&The top Democrat on the Senate Finance Committee asked the Trump administration on Thursday to turn over the names of visitors $\ldots$. That investigation found that members of the golf clubs Trump visited most often as president $\dots$. Membership lists at Trump's private clubs are secret. USA TODAY found the names of about 4,500 members by reviewing $\ldots$ . In a letter to the Department of Homeland Security's Acting Secretary Elaine Duke, Wyden said USA TODAY's examination \dots\\
\Tstrut &\multicolumn{1}{c}{\bf Ground-Truth}\\
\hline
\Tstrut Headline & Senator seeks visitor logs, golf partners
\\
\mbox{Highlight} & An investigation by USA TODAY prompts a senior Democratic senator to seek visitor logs at Trump clubs and names of his golfing partners .\\
Teaser & \boldblue{Citing} USA TODAY’s investigation, a top \boldblue{Sen.} Democrat \boldblue{seeks} visitor \boldblue{logs} to Trump’s golf courses \& golfing partners.\\
\Tstrut &\multicolumn{1}{c}{\bf Generated}\\
\hline
\Tstrut Headline & trump [UNK] to turn over trump 's private clubs
\\
\mbox{Highlight} & the top democrat on the senate finance committee asked the trump administration .\\
Teaser & the top democrat on trump 's golf club : `` it ’ s a \boldblue{lot} of \boldblue{money} , but it ’ s not \boldblue{going} to\\
\hline
\end{tabular}
}
\end{center}
\caption{seq2seq\_pointer generated examples. Non-overlapping words are in bold blue.}
\end{table} 

\end{document}